\documentclass[conference]{IEEEtran}
\IEEEoverridecommandlockouts
\usepackage{cite}
\usepackage{amsmath,amssymb,amsfonts}
\usepackage{algorithmic}
\usepackage{graphicx}
\usepackage{textcomp}
\usepackage{xcolor}
\def\BibTeX{{\rm B\kern-.05em{\sc i\kern-.025em b}\kern-.08em
    T\kern-.1667em\lower.7ex\hbox{E}\kern-.125emX}}

\usepackage{subfig} 
\usepackage{booktabs} 
\usepackage{makecell} 

\newcommand{\comment}[1]{ }

\newcommand{\wkw}[1]{\textcolor{red}{[WKW: #1]}}

\begin{document}
\bstctlcite{IEEEexample:BSTcontrol}



\title{Unsupervised Contrastive Learning for Robust RF Device Fingerprinting Under Time-Domain Shift}

\author{Jun Chen, Weng-Keen Wong, Bechir Hamdaoui\\
Oregon State University, Corvallis, OR, USA\\
Email: \{chenju3,wongwe,hamdaoui\}@oregonstate.edu \\
}



\maketitle
\thispagestyle{plain}
\pagestyle{plain}

\begin{abstract}
Radio Frequency (RF) device fingerprinting has been recognized as a potential technology for enabling automated wireless device identification and classification. However, it faces a key challenge due to the domain shift that could arise from variations in the channel conditions and environmental settings, potentially degrading the accuracy of RF-based device classification when testing and training data is collected in different domains. 
This paper introduces a novel solution that leverages contrastive learning to mitigate this domain shift problem.
Contrastive learning, a state-of-the-art self-supervised learning approach from deep learning, learns a distance metric such that positive pairs are closer (i.e. more similar) in the learned metric space than negative pairs. When applied to RF fingerprinting, our model treats RF signals from the same transmission as positive pairs and those from different transmissions as negative pairs. Through experiments on wireless and wired RF datasets collected over several days, we demonstrate that our contrastive learning approach captures domain-invariant features, diminishing the effects of domain-specific variations. Our results show large and consistent improvements in accuracy (10.8\% to 27.8\%) over baseline models, thus underscoring the effectiveness of contrastive learning in improving device classification under domain shift.
%

\footnote{This work is supported in part by NSF/Intel Award No. 2003273.}
\footnote{\textregistered 2024 IEEE. Personal use of this material is permitted. Permission from IEEE must be obtained for all other uses, in any current or future media, including reprinting/republishing this material for advertising or promotional purposes, creating new collective works, for resale or redistribution to servers or lists, or reuse of any copyrighted component of this work in other works.}
\end{abstract}


\section{Introduction}
\label{sec:intro}
Radio frequency (RF) device fingerprinting~\cite{Soltanieh2020ARO} plays an important role in network security, enabling physical-layer-based network access authentication and network device classification through its ability to identify devices from their transmitted RF signals. The use of deep learning~\cite{Jian2020DeepLF,hamdaoui2023deep,elmaghbub2021lora,hamdaoui2020deep,
AlShawabka2020ExposingTF,puppo2023hinova} in RF fingerprinting has become prevalent in recent years as it enabled the extraction of device-specific features and signatures solely from sampled raw RF signals, thereby eliminating the need for data preprocessing and domain knowledge. 


A significant real-world issue that affects the accuracy of deep learning models is domain shift~\cite{hamdaoui2022uncovering,Zhao2019OnLI}, which occurs when the training data from the \emph{source} domain has a different distribution than the distribution of the test data (i.e. the data encountered during deployment) in the \emph{target} domain.  This mismatch hampers the performance as models trained on the source domain struggle to adapt to the characteristics of the new target domain encountered during deployment, leading to a degradation in performance~\cite{AlShawabka2020ExposingTF,elmaghbub2021lora,hamdaoui2022uncovering}. 

Much attention has been given to the area of domain adaptation (DA) in machine learning, which has the goal of dealing with domain shift. Success in domain adaptation hinges on generalizing knowledge from the source to the target domain by identifying domain-invariant aspects of the data or modeling the differences between the source and target domains. Various categories of deep learning-based DA techniques exist. For instance, alignment-based methods \cite{Sun2016DeepCC} concentrate on aligning feature distributions between source and target domains. Adversarial-based methods \cite{Ganin2015DomainAdversarialTO} employ a domain discriminator to differentiate domains and a feature extractor to create indistinguishable features. Finally, disentanglement-based methods \cite{Bousmalis2016DomainSN,Elmaghbub2023ADLIDAD} learn to separate domain-specific and domain-invariant features.

\comment{
\begin{figure}
    \centering 
    \subfloat[Domain Adaptation]{%
      \includegraphics[width=0.235\textwidth]{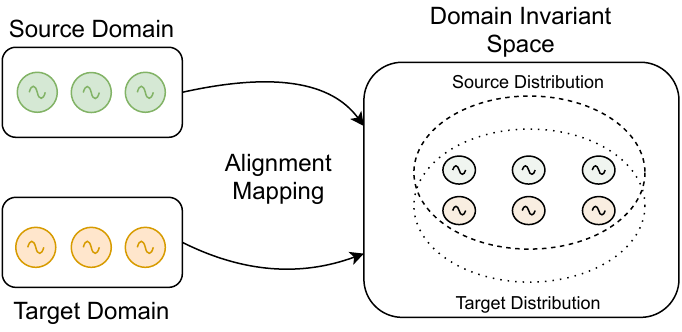}
      \label{fig:illus_da}} \hspace*{-0.2em}
    \subfloat[Contrastive Learning]{%
      \includegraphics[width=0.245\textwidth]{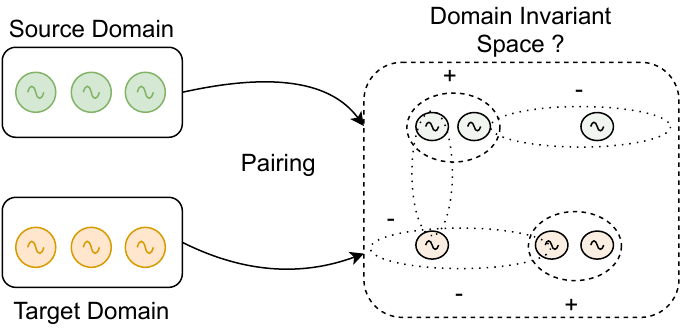}
      \label{fig:illus_cl}} \hspace*{-0.2em}
    \caption{Conceptual diagrams for domain adaptation and contrastive learning. (a) Domain adaptation aims to learn domain-invariant features by aligning data distribution from different domains. (b) Contrastive learning aims to learn interior features by contrasting positive pairs with negative pairs. \wkw{This diagram is confusing to follow and more importantly, it takes up space to try to explain something that isn't all that necessary to explain. I recommend not including it. We can explain things more clearly in the contrastive learning paragraph below.}}
    \label{fig:illus_da_cl}
\end{figure}
}

Dealing with domain shift in RF fingerprinting is crucial for ensuring accurate and reliable device identification and classification. Unfortunately, most recent deep learning domain adaptation techniques~\cite{Bousmalis2016DomainSN, Sun2016DeepCC,Wilson2018ASO} are designed for computer images and are not readily applicable to the complexities of RF data. Furthermore, domain adaptation is especially challenging in RF fingerprinting, as RF signals exhibit domain shift from many different factors such as non-stationarity and variation due to changes in  environmental/network settings. These variations introduce domain-specific characteristics into RF fingerprints, posing a challenge in developing accurate and robust fingerprinting models that can generalize across different domains.

A handful of approaches have been proposed to mitigate the effects of domain shift in RF device fingerprinting models. For instance, in \cite{wang2021specific,wang2022specific},  multi-discrepancy and adversarial methods have been employed to address the domain adaptation classification of RF device fingerprinting under different SNRs using both wired and simulated Rayleigh channel models. A disentanglement-based method is also applied to address the domain adaptation problem in RF device fingerprinting \cite{Elmaghbub2023ADLIDAD}, which aims to separate the domain-specific and domain-invariant features present in the data. 


In contrast to previous approaches, we introduce a domain adaptation method that avoids the complexities associated with alignment techniques, the need for strong domain knowledge assumptions, or the instability often seen in adversarial training. Our approach is based on contrastive learning \cite{Noroozi2016UnsupervisedLO, He2019MomentumCF, Chen2020ImprovedBW, Chen2021AnES}, which has been shown to be a highly effective self-supervised learning approach \cite{Balestriero2023ACO} for computer vision and natural language processing.  Contrastive learning  relies on a pretext task during a pre-training phase. This pretext task is designed to train the network to learn an informative representation such that the learned representation can be useful for a downstream task such as classification. During pre-training, the network optimizes a contrastive loss that results in a distance metric such that a pair of similar instances (called a positive pair) is closer together in this metric space than a pair of dissimilar instances (called a negative pair). The ability of contrastive learning to identify salient features from positive and negative pairs makes it a natural fit for domain adaptation.


Contrastive learning has been extensively applied to images and video~\cite{Noroozi2016UnsupervisedLO,He2019MomentumCF,Grill2020BootstrapYO,Chen2020ExploringSS,Chen2021AnES} but much less attention has been given to time series data like RF signal data emitted by RF wireless devices \cite{ozyurt2023contrastive}. Recent work applies contrastive learning to domain adaptation in the context of computer vision ~\cite{Thota2021ContrastiveDA}, but to our knowledge, our work represents the first application of contrastive learning to address domain adaptation in RF data-based device fingerprinting and classification.
When we apply contrastive learning to RF device fingerprinting, we rely on a fundamental assumption: data from the same transmission form positive pairs while data from different transmissions form negative pairs. Our research shows that by incorporating this single and intuitive assumption, the contrastive learning framework guides the model to emphasize the discriminative aspects of RF signals while effectively ignoring domain-specific variations caused by factors like channel conditions and noise.

The rest of this paper is organized as follows. Section II describes the proposed method in detail. Section III presents the experimental results and analysis. Finally, Section IV concludes the paper and discusses future work.


\section{Methodology}
\label{sec:methodlogy}

To make the trained model adaptable to the target domain, we leverage contrastive learning to allow the model to learn domain-invariant representations that transfer across domains, resulting in better performance and generalization. In the following sections, we provide details on the dataset construction, as well as on the pre-training, training and testing phases of the proposed constrastive learning based framework.

\subsection{WiFi Testbed and RF Dataset Collection} \label{sec:testset}

We used real RF datasets collected using our experimental testbed of $15+$ PyCom/IoT transmitting devices (a combination of Fipy and Lopy boards). 
The Pycom devices were programmed to transmit IEEE 802.11b WiFi frames at a center frequency of $2.412$GHz and a bandwidth of $20$MHz. 
These transmitted frames were sampled and stored as IQ (In-phase and Quadrature) data by an
Ettus USRP (Universal Software Radio Peripheral) B$210$ receiver at a sample rate of $45$MSps. Each WiFi capture lasts for $2$ minutes generating more than $5000$ frames per device with each frame consisting of $25170$ complex-valued samples. 
We experimented with two scenarios: Wireless-WiFi and Wired-WiFi. 
In the wireless scenario, the transmitters were all located $1$m away from the receiver and transmitted wirelessly, whereas in the wired scenario, the transmitters were directly connected to the receiver through an SMA cable, allowing to mask the impact the wireless channel. More details on the testbed can be found in~\cite{hamdaoui2022deep,johnson2023domain,elmaghbub2023eps}.

\subsection{Dataset Construction} \label{sec:dc}

\begin{figure}
    \centering
    \includegraphics[width=0.45\textwidth]{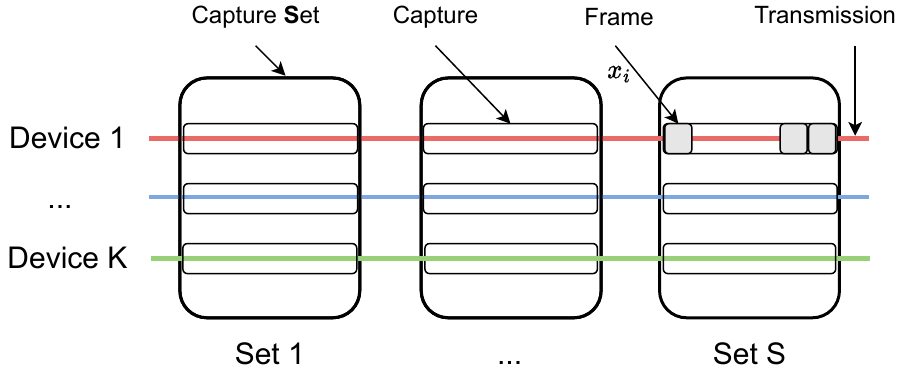}
    \caption{An overview of how training and test data are created.}
    \label{fig:ds_split}
\end{figure}

Fig.~\ref{fig:ds_split} illustrates the process for creating training and test data using WiFi transmissions from the $K=15$ different Pycom devices described in the previous section. Specifically, we created $S$ datasets---referred to as \emph{Capture Sets or Sets}, which consist of portions of the tested devices' transmissions all captured at the same offset in time from the start of the transmission. 
To ensure that the training sets and test sets experience some form of domain shift, we deliberately choose Sets captured on different days; one day for the source domain and another day for the target domain. For example, we chose \emph{Set 1} of day 1 (denoted as D1S1) as the source domain and \emph{Set 1} of day 2 (denoted as D2S1) as the target domain. 


Within a \emph{Set}, we refer to the data belonging to a \emph{single} device as a \emph{Capture}, which consists of $2500$ consecutive, non-overlapping IQ sliding windows (referred to as \emph{frames}) with windows size of $1000$. Each \emph{Set} is formed by taking the union of the IQ frames from a transmission over all $15$ devices, resulting in $2500\times 15$ frames in total within the \emph{Set}. Each frame has dimensions of $2\times 1000$ because the IQ sample has the two dimensions of $I$ and $Q$. 

\subsection{The Contrastive-Based Domain Adaptation Framework} \label{sec:tp}



\begin{figure*}
    \centering
    \begin{minipage}[c]{0.60\textwidth}
        \centering
        \subfloat[Pre-training Stage]{%
            \includegraphics[width=\textwidth]{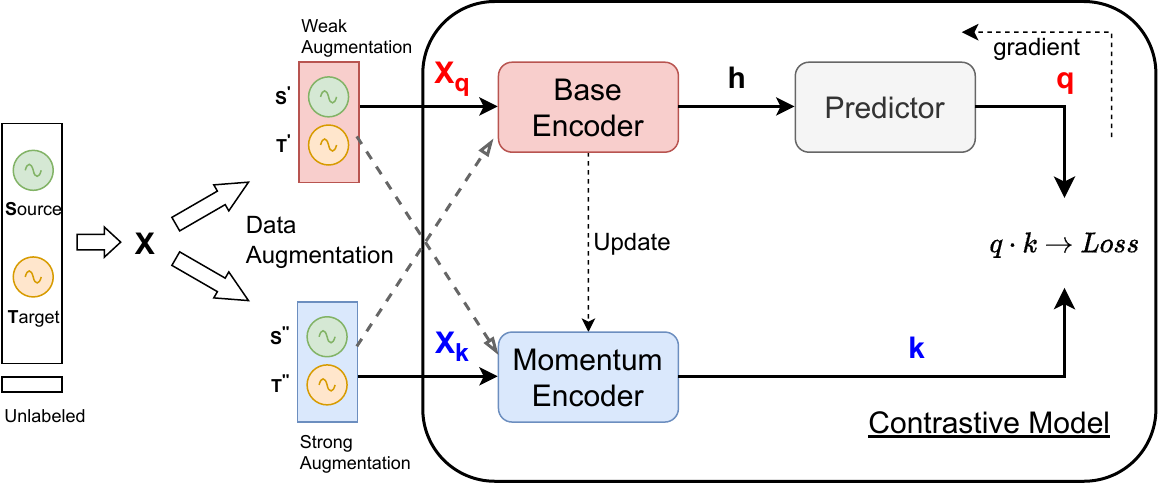}
            \label{fig:train_phase1}} 
    \end{minipage}
    \hfill
    \begin{minipage}[c]{0.35\textwidth}
        \centering
        \subfloat[Training Stage]{%
            \includegraphics[width=\textwidth]{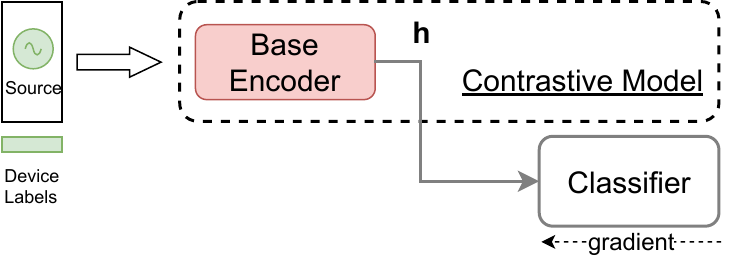}
            \label{fig:train_phase2}} 
        \vspace{0.5em}
        \subfloat[Testing Stage]{%
            \includegraphics[width=\textwidth]{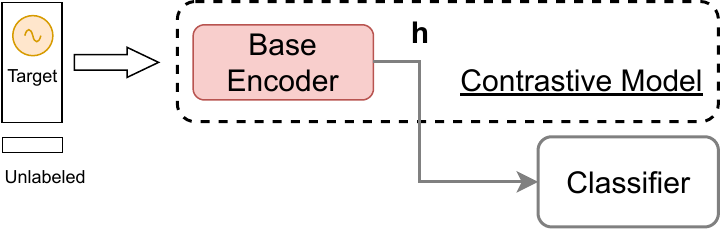}
            \label{fig:train_phase3}} 
    \end{minipage}
    \caption{The proposed training procedure and contrastive-based framework for addressing the domain shift issue in RF fingerprinting. The overall domain adaptation process includes three stages: (a) pre-training, (b) training and (c) testing. The source inputs consist of IQ frames from one day, while the target inputs are from another day. The pre-training stage uses unlabeled data, meaning we do not know which device produced the data but we do know which captures come from the same transmission. Labeled data in the form of device labels is only used during the training stage to train the classifier.} 
    
    \label{fig:train_phase}
\end{figure*}
Fig.~\ref{fig:train_phase} illustrates the three distinct stages of the proposed domain adaptation framework: the pre-training stage, the training stage and the testing stage. In the first stage, we pre-train the model based on the source and target dataset with a contrastive learning approach. The goal of the pre-training stage is to learn a domain-invariant representation without device labels through the use of a base encoder. In the second stage, we use the base encoder as a feature extractor and train a classifier based on the extracted features and the device labels from the source domain only. Finally, we apply the trained base encoder and the trained classifier on the target data during the testing stage.
Next, we provide detailed description of each of the three stages.

\subsubsection{Pre-training Stage} \label{sec:pretr}



We now describe the neural network framework used during the pre-training stage, as well as during the training and testing stage (i.e., the classifier). During pre-training, we use a model that is based on MoCo V3 \cite{Chen2021AnES} with a modified ResNet-18 \cite{He2015DeepRL} network as the internal base encoder. For the training stage, the classifier is a fully connected neural network with two hidden layers. Finally, for the testing stage, the classifier has the same architecture as in the training phase. The detailed components in the pre-training framework are described below:

\paragraph{Data Augmentation} \label{sec:daug}


\begin{figure}
    \centering 
    \subfloat[Weak Augmentation]{%
      \includegraphics[width=0.24\textwidth]{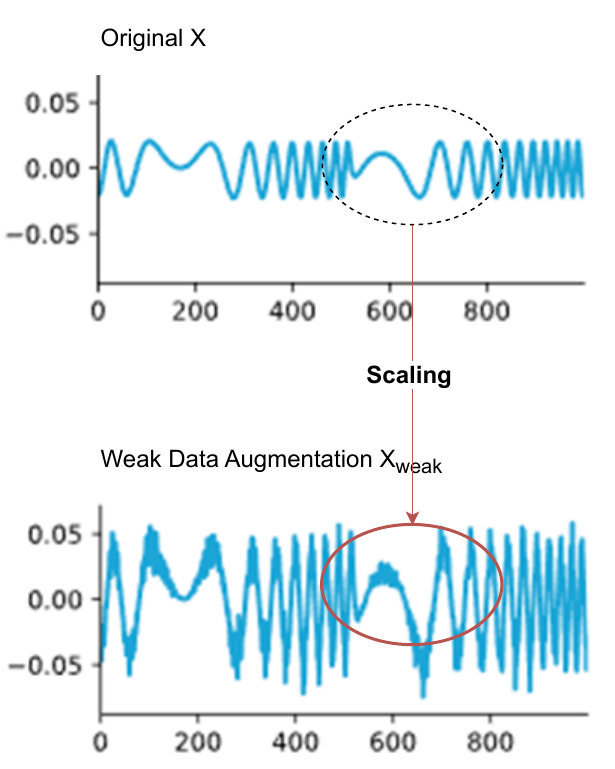}
      \label{fig:ts_aug_owk}} \hspace*{-0.5em}
    \subfloat[Strong Augmentation]{%
      \includegraphics[width=0.24\textwidth]{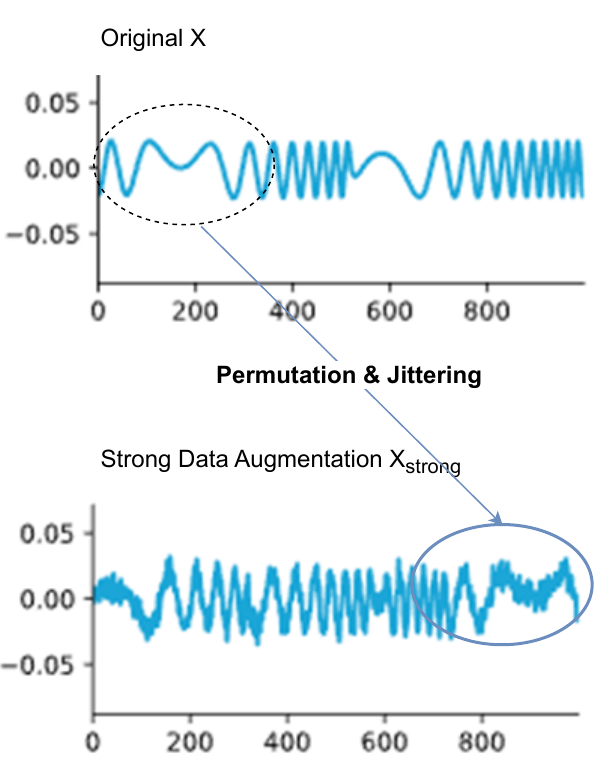}
      \label{fig:ts_aug_ost}} \hspace*{-0.5em}
    \caption{Illustrations of weak and strong data augmentation.}
    \label{fig:ts_aug}
  \end{figure}


A data augmentation process is needed to generate positive and negative pairs for contrastive learning by creating transformations from the original training data. While several well-known data augmentations for contrastive learning have been proposed in the literature (e.g. rotation and flipping), many are not applicable to our RF fingerprinting domain. Following \cite{Um2017DataAO}, we employ both weak and strong data augmentations for time series. Weak augmentation (Fig. \ref{fig:ts_aug_owk}) includes random scaling of IQ data, meaning the augmented data is obtained by multiplying the original data by a scaling factor. Strong augmentation (Fig. \ref{fig:ts_aug_ost}) involves random jittering and segmentation permutation of IQ data. Jittering entails a random amplitude shift of the IQ data, achieved by adding random noise to the original data, while segmentation permutation is a random rearrangement of intervals of the IQ data along the time axis. This augmentation enhances the model's resilience to amplitude shifts and time permutations in the IQ data.

For any data frame $X$, there are two augmented data frames produced ($X_{weak}$ and $X_{strong}$ as shown in Fig.~\ref{fig:ts_aug}), which are fed into the contrastive model in Fig.~\ref{fig:train_phase1} as query vector $X_q$ (the input of the base encoder in the contrastive model) and key vector $X_k$ (the input of the momentum encoder) respectively. Intuitively, our contrastive learning model acts as a dictionary lookup, which will match the query vector $X_q$ (identifier) with the key vector $X_k$ (vector retrieved from the batch) as a positive pair and pull them closer. Fig.~\ref{fig:train_phase1} shows how IQ data is augmented and fed into the contrastive model. The solid arrow line represents feeding $X_{weak}$ to the base encoder as $X_q$; similarly, $X_{strong}$ is fed to the momentum encoder as $X_k$. In addition, the dashed arrow lines also send $X_{strong}$ to the base encoder and $X_{weak}$ to the momentum encoder to ensure symmetry in the loss function.

\paragraph{Pre-train Model} \label{sec:model}

The pre-train model (Fig. \ref{fig:train_phase1}) consists of two key encoders: the base encoder, with output denoted as $h$, and the momentum encoder, with output denoted as $k$. The base encoder uses a modified ResNet-18 architecture as the backbone, where the modifications include adjustments to the first convolutional layer (kernel size 100, stride size 20) and the last convolutional layer (outputs a 128-dimensional vector). The last few layers of the base encoder are a multi-layer perceptron (MLP) projector network, which effectively replaces ResNet-18's final fully connected layer. This projector network produces a 128-dimensional feature vector for use by the classifier during training. During pre-training, this output vector $h$ is processed by a MLP predictor network, generating an output vector $q$ for the contrastive loss function. The momentum encoder mirrors the base encoder's architecture and the same momentum update done in MOCO \cite{He2019MomentumCF} keeps the key representation up-to-date.

\paragraph{Loss Function} \label{sec:loss}
The pretext task of the contrastive model is to predict $q$ such that it is close in distance to $k$ by minimizing the modified soft nearest neighbor loss \cite{Frosst2019AnalyzingAI} shown in Eq.~\eqref{eq:snn_loss}. This loss is computed over a batch of $N$ frames. In Eq.~\eqref{eq:snn_loss}, $q_i$ is the output of the predictor for the $i$-th frame $X_i$, $k_j$ is the output of the momentum encoder for the $j$th frame in the batch and $y$ is the transmission ID. To understand the soft nearest neighbor loss function, consider the fraction inside the log term. For a given $q_i$, this fraction represents the probability of selecting a frame $k_j$ such that it comes from the same transmission as $q_i$. This probability depends on the distance between $k_j$ and $q_i$, where the distance is measured as an exponentiated negative Euclidean distance between $q_i$ and $k_j$, along with a temperature hyperparameter $\tau$ (which we set to 0.2) that adjusts the weight of these distances. Putting everything together, the soft nearest neighbor loss is the average negative log probability of selecting $k_j$ such that it comes from the same transmission as $q_i$ (denoted as the transmission ID $y_{q_i}=y_{k_j}$).
Thus, minimizing the loss function reduces the distance, in the learned representation space, between the $i$-th frame and all the frames from the same transmission.


\begin{equation}
    \begin{aligned}
         & L_{snn}(q, k, y, \tau)=                                                                      \\
         & -\frac{1}{N} \sum_{i \in 1 ... N} \log \left(\frac{\sum_{\substack{j \in 1 ... N             \\
        y_{q_{i}}=y_{k_{j}}}} \exp{(-\frac{\left\|q_i-k_j\right\|^2}{\tau})}}{\sum_{\substack{k \in 1 ... N \\
        }} \exp{(-\frac{\left\|q_i-k_k\right\|^2}{\tau})}}\right)                                           \\
         &
        \label{eq:snn_loss}
    \end{aligned}
\end{equation}


Following the example of MoCo V3 \cite{Chen2021AnES}, both $X_{strong}$ and $X_{weak}$ should possess the capability to predict each other and we symmmetrize the contrastive loss function to be:
$L_{ct} = 2 \tau [L_{snn}(q_w, k_s, y, \tau) + L_{snn}(q_s, k_w, y, \tau)]$. 
In this symmetric version, we denote the output of the predictor as $q_w$ and output of momentum encoder as $k_s$ for the case of sending $X_{weak}$ into the base encoder and $X_{strong}$ into the momentum encoder. Similarly, for its symmetric counterpart, we denote the output of the predictor as $q_s$ and output of momentum encoder as $k_w$ for the case of sending $X_{strong}$ into base encoder and $X_{weak}$ into momentum encoder. 

\comment{
\begin{equation}
    L_{ct} = 2 \tau [L_{snn}(q_w, k_s, y, \tau) + L_{snn}(q_s, k_w, y, \tau)]
    \label{eq:ct_loss}
\end{equation}
}


\subsubsection{Training Stage} \label{sec:clftr}

After the pre-training stage, we will use the learned base encoder as the feature extractor to produce a representation of the training data that is fed to the classifier. We will train the classifier using labeled data (i.e. data that has device labels) from the source domain (see Fig.~\ref{fig:train_phase2}) so that it can identify individual devices.


The classifier is a fully connected neural network with two hidden layers. The input of the classifier is a 128-dimensional vector extracted by the base encoder of pre-trained model, and the output vector has a dimension of the number of devices, which is the number of classes in the source and target domain. The classifier is trained based on the cross-entropy loss function.

\subsubsection{Testing Stage} \label{sec:clftt}
In this stage, the classifier predicts the device ID using input frames from the target domain (see Fig.~\ref{fig:train_phase3}). Frames are first fed into the base encoder to produce a new feature vector corresponding to a domain-invariant representation of the input frame. This new feature vector is provided as input to the classifier, which then predicts the device label.

\section{Results}
\label{sec:results}



We present results that compare our contrastive learning model (CTL) against two baseline models. The first baseline model (CNN) is a CNN classifier based on a ResNet18 backbone. The second baseline model (AB) is an ablated variant of our contrastive learning model which has the same architecture as the contrastive learning model, but it only uses the source domain data for pre-training. In contrast, the CTL model uses both source and target domain data.

\begin{figure*}
  \centering 
  \subfloat[Wired, Day 1 $\leftrightarrow$ 2]{%
    \includegraphics[width=0.33\textwidth]{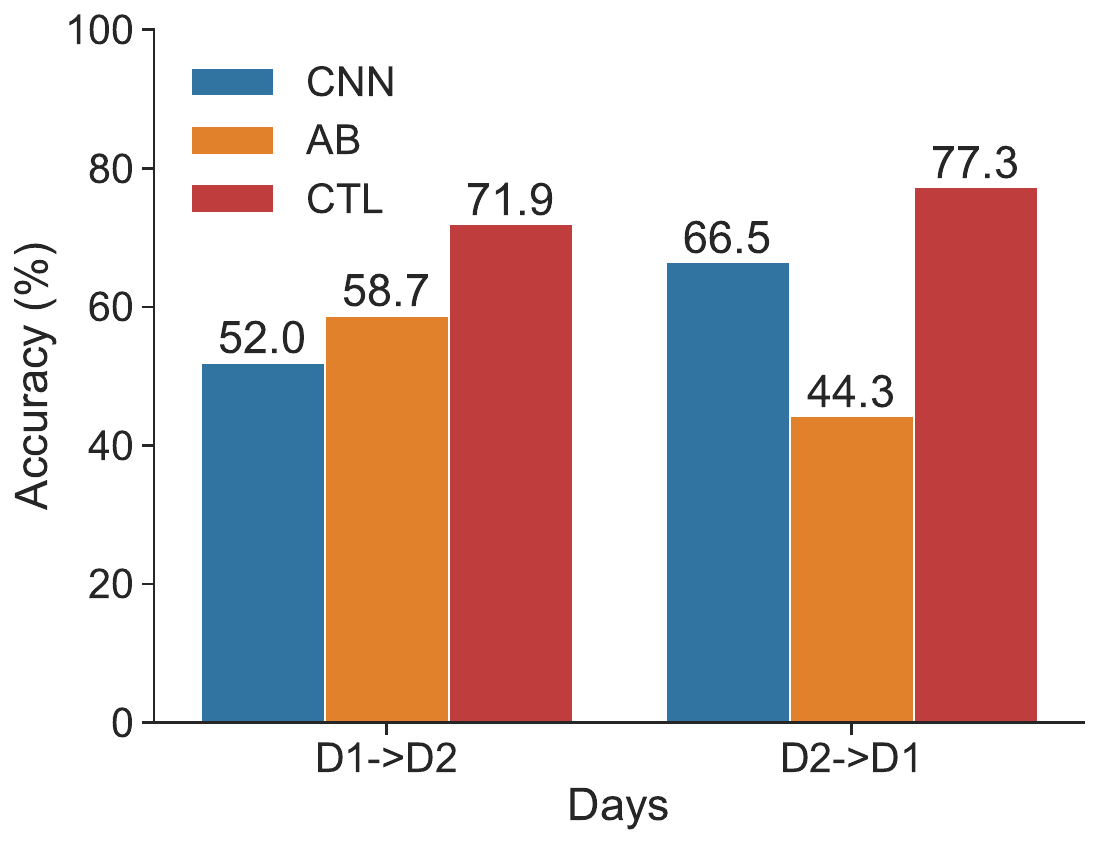}
    \label{fig:expr_bar_wired12}} \hspace*{-0.5em}
  \subfloat[Wireless, Day 1 $\leftrightarrow$ 2]{%
    \includegraphics[width=0.33\textwidth]{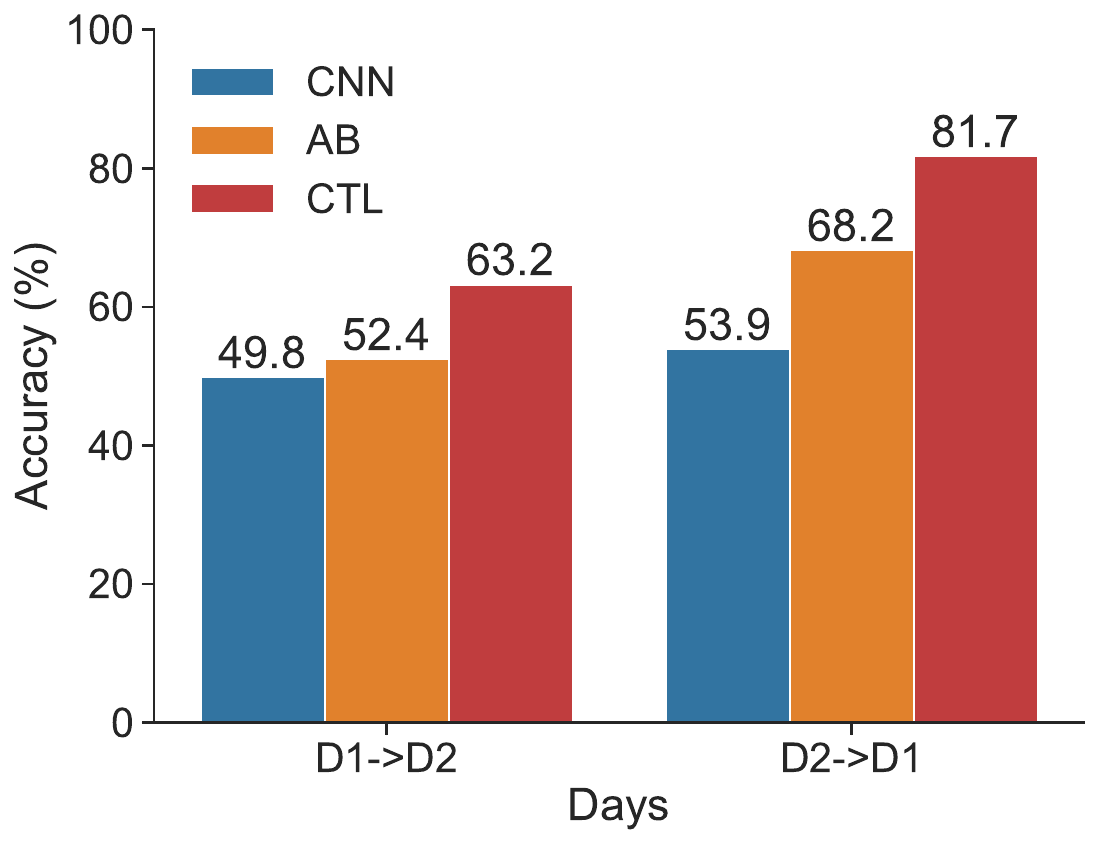}
    \label{fig:expr_bar_wireless12}} \hspace*{-0.5em}
  \subfloat[Wireless, Day 2 $\leftrightarrow$ 3]{%
    \includegraphics[width=0.33\textwidth]{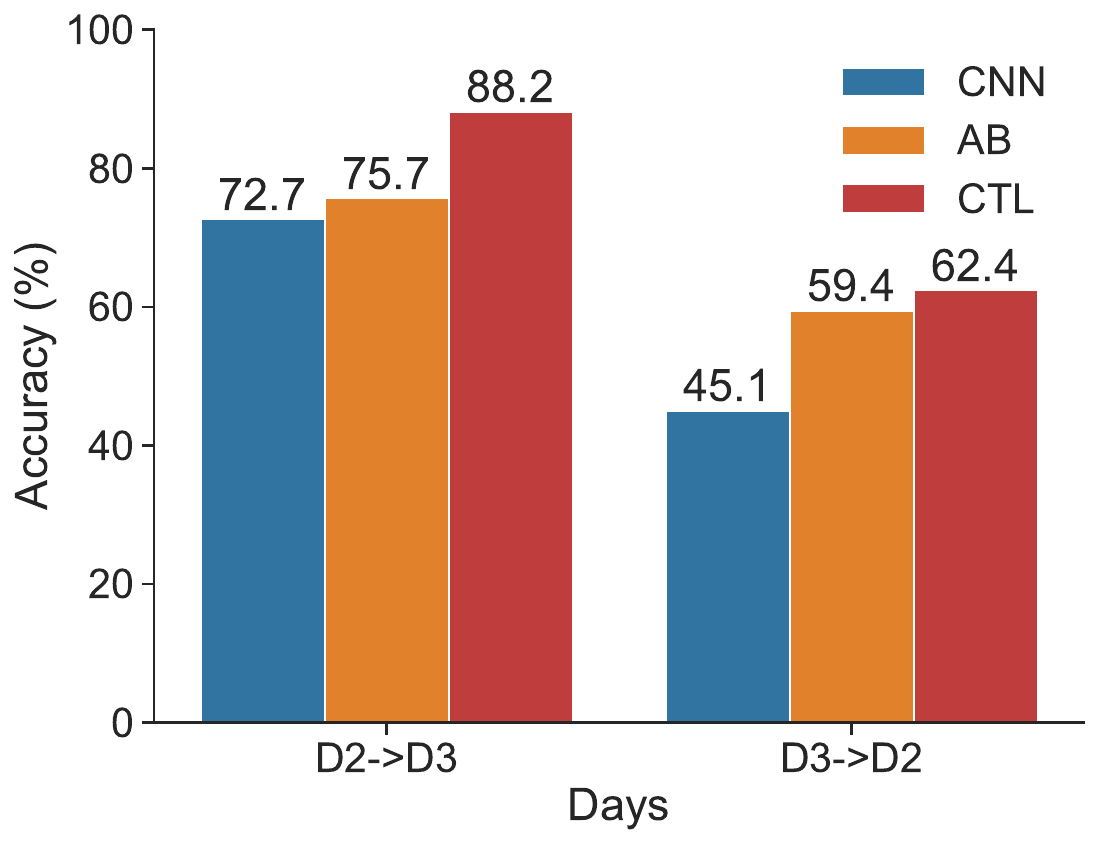}
    \label{fig:expr_bar_wireless23}} \hspace*{-0.5em}
  \caption{Domain adaptation accuracy: from one day to another day on wired and wireless RF devices for CNN, AB and CTL.}
  \label{fig:expr_bar_all}
\end{figure*}

We considered wired and wireless scenarios, each consisting of several Capture Sets from the same transmission (refer to Section~\ref{sec:dc} for more details). Fig.~\ref{fig:expr_bar_all} reports results based on the average over all \emph{Sets}, while Tables~\ref{tab:expr_tb_wired} to \ref{tab:expr_tb_wireless23} report the results of each individual \emph{Set}.

\subsection{Classification Accuracy: Comparison with CNN} \label{sec:clfds}



The CTL model produces large increases in accuracy (10.8-19.9\%) over the CNN baseline in wired scenarios (Fig.~\ref{fig:expr_bar_wired12}). Table~\ref{tab:expr_tb_wired} shows the CTL model consistently outperforming the CNN model across all \emph{Sets} of the wired RF data, with accuracy improvements of 9.9-24.8\% on individual \emph{Sets}.


\begin{table}[!htb]
  \centering \resizebox{0.95\columnwidth}{!}{
    \begin{tabular}{@{}l|lll|lll@{}}
      \toprule
      \multicolumn{1}{l|}{\thead{Wired (Source $\rightarrow$ Target)}} & \multicolumn{3}{l|}{\thead{Day 1 $\rightarrow$ 2}} & \multicolumn{3}{l}{\thead{Day 2 $\rightarrow$ 1}} \\
      \toprule
      \#Device K = 16 & CNN & AB & CTL & CNN & AB & CTL \\
      \midrule
      DayA\_S1  $\rightarrow$ DayB\_S1 & 50.4 & 57.8 & 71.9 & 67.5 & 46.0 & 78.8\\
      DayA\_S1  $\rightarrow$ DayB\_S2 & 51.6 & 58.2 & 71.0 & 65.1 & 45.4 & 75.4\\
      DayA\_S1  $\rightarrow$ DayB\_S3 & 53.6 & 57.9 & 67.6 & 66.5 & 42.8 & 76.4\\
      DayA\_S1  $\rightarrow$ DayB\_S4 & 52.4 & 60.9 & 77.2 & 66.9 & 42.8 & 78.6\\
      \bottomrule
    \end{tabular}
  }
  \caption{Classification accuracy of domain adaptation between individual \emph{Sets} of day 1 and day 2 on \textbf{wired} RF devices for CNN, AB and CTL.}
  \label{tab:expr_tb_wired}
\end{table}





Fig.~\ref{fig:expr_bar_wireless12} and \ref{fig:expr_bar_wireless23} show that in wireless scenarios, contrastive learning once again substantially improves accuracy (13.4\%-27.8\%) over the CNN baseline. Tables~\ref{tab:expr_tb_wireless12} and \ref{tab:expr_tb_wireless23} show that CTL consistently produces large increases in accuracy (7.6\%-34.1\%) for individual \emph{Sets}. 

\begin{table}[!htb]
  \centering \resizebox{0.95\columnwidth}{!}{
    \begin{tabular}{@{}l|lll|lll@{}}
      \toprule
      \multicolumn{1}{l|}{\thead{Wireless (Source $\rightarrow$ Target)}} & \multicolumn{3}{l|}{\thead{Day 1 $\rightarrow$ 2}} & \multicolumn{3}{l}{\thead{Day 2 $\rightarrow$ 1}} \\
      \toprule
      \#Device K=15 & CNN & AB & CTL & CNN & AB & CTL \\
      \midrule
      DayA\_S1  $\rightarrow$ DayB\_S1 & 35.4 & 39.6 & 53.5 & 42.3 & 65.9 & 68.5\\
      DayA\_S1  $\rightarrow$ DayB\_S2 & 58.3 & 62.2 & 67.1 & 54.1 & 66.6 & 86.2\\
      DayA\_S1  $\rightarrow$ DayB\_S3 & 52.8 & 55.1 & 66.9 & 60.5 & 70.4 & 83.0\\
      DayA\_S1  $\rightarrow$ DayB\_S4 & 52.8 & 52.8 & 65.3 & 58.7 & 70.0 & 89.2\\
      \bottomrule
    \end{tabular}
  }
  \caption{Detailed classification accuracy of domain adaptation between one \emph{Set} of day 1 and another \emph{Set} of day 2 on \textbf{wireless} RF devices for CNN, AB and CTL.}
  \label{tab:expr_tb_wireless12}
\end{table}


\begin{table}[!htb]
  \centering \resizebox{0.95\columnwidth}{!}{
    \begin{tabular}{@{}l|lll|lll@{}}
      \toprule
      \multicolumn{1}{l|}{\thead{Wireless (Source $\rightarrow$ Target)}} & \multicolumn{3}{l|}{\thead{Day 2 $\rightarrow$ 3}} & \multicolumn{3}{l}{\thead{Day 3 $\rightarrow$ 2}} \\
      \toprule
      \#Device K=15 & CNN & AB & CTL & CNN & AB & CTL \\
      \midrule
      DayA\_S1  $\rightarrow$ DayB\_S1 & 61.7 & 63.0 & 71.1 & 33.8 & 64.0 & 67.9\\
      DayA\_S1  $\rightarrow$ DayB\_S2 & 76.3 & 84.3 & 94.6 & 53.4 & 61.3 & 71.0\\
      DayA\_S1  $\rightarrow$ DayB\_S3 & 77.3 & 79.3 & 94.5 & 46.5 & 54.7 & 56.9\\
      DayA\_S1  $\rightarrow$ DayB\_S4 & 75.2 & 76.2 & 92.6 & 46.4 & 57.7 & 54.0\\
      \bottomrule
    \end{tabular}
  }
  \caption{Detailed classification accuracy of domain adaptation between one \emph{Set} of day 2 and another \emph{Set} of day 3 on \textbf{wireless} RF devices for CNN, AB and CTL.}
  \label{tab:expr_tb_wireless23}
\end{table}



\subsection{Classification Accuracy: Comparison with AB} \label{sec:ab}
Fig.~\ref{fig:expr_bar_all} shows that the CTL model outperforms the AB model, resulting in increases in accuracy of 13.2\%-33.0\% for wired and 3.0\%-13.5\% for wireless. The CTL model outperforms AB in all individual sets in both scenarios, as shown in Tables~\ref{tab:expr_tb_wireless12} and \ref{tab:expr_tb_wireless23}. The results of the AB model highlight the importance of including target domain data in pre-training. We emphasize the fact that the target domain data is unlabeled, meaning that it does not contain device labels identifying the device that generate a frame; without device labels, the CTL model is not given any information that matches frames in the source domain with frames in the target domain generated by the same device. The high accuracy of the CTL model indicates that even without explicit device labels in the target domain, simply knowing that certain frames come from the same transmission enables the CTL model to capture crucial domain-invariant features that greatly improve its adaptability and generalization.



For the majority of the experiments, the AB model outperforms the CNN model, highlighting the importance of pre-training with source data. However, an exception is observed in the "Day 2 $\rightarrow$ 1" case of Fig.~\ref{fig:expr_bar_wired12}, where the AB model's accuracy (44.3\%) is much lower than that of the CNN model (66.5\%). This result suggests that pre-training with only source domain data risks capturing domain-specific information, which could harm performance. The CTL model, however, combines target domain data with source domain data in the pre-training process, thus capturing essential domain-invariant features crucial for successful domain adaptation.

\subsection{Confusion Matrix View on Device Level Analysis} \label{sec:cma}

\begin{figure}
  \centering 
  \subfloat[CNN, Day 1 $\rightarrow$ 2]{%
    \includegraphics[width=0.22\textwidth]{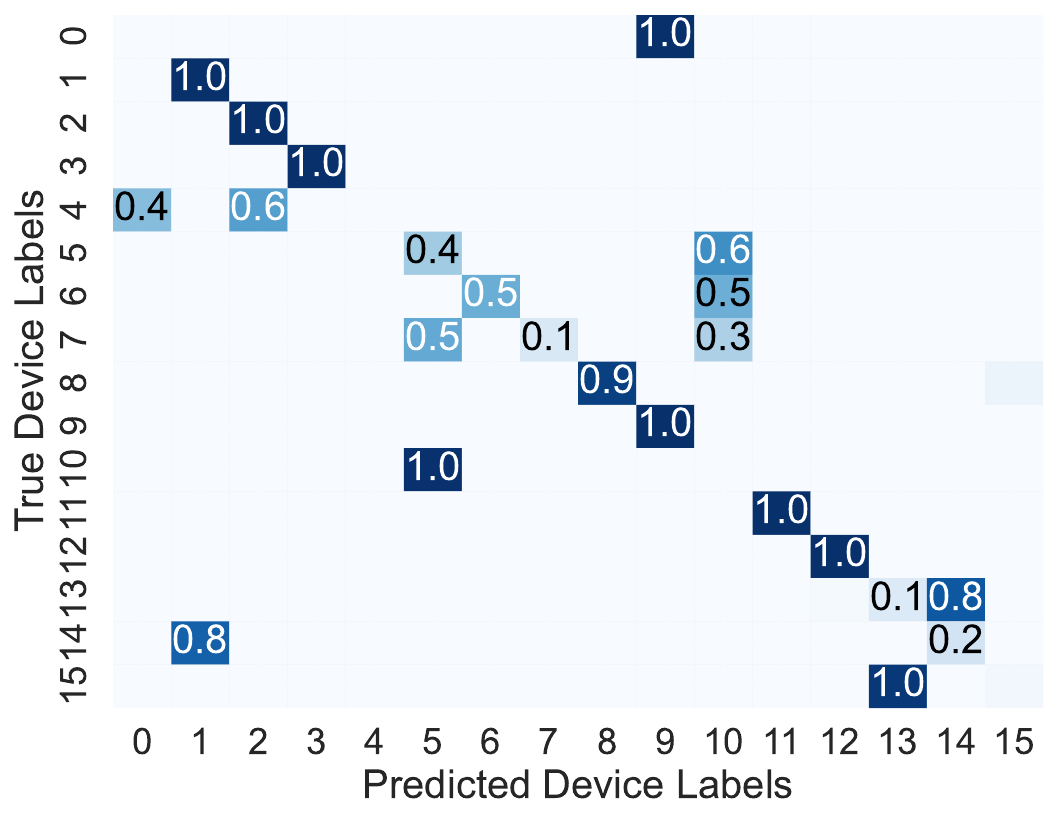}
    \label{fig:expr_cm_cnn_wired12x}} \hspace*{-0.1em} 
  \subfloat[CNN, Day 2 $\rightarrow$ 1]{%
    \includegraphics[width=0.22\textwidth]{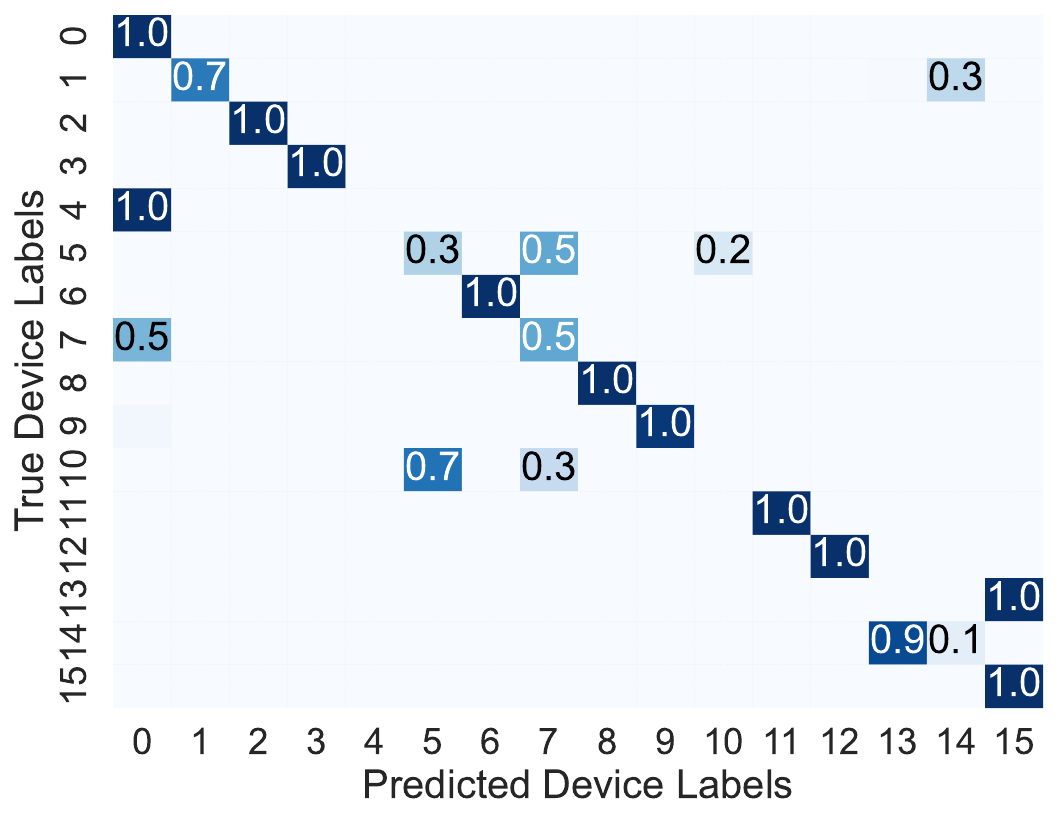}
    \label{fig:expr_cm_cnn_wired21x}} \hspace*{-0.1em}
  \newline
  \subfloat[AB, Day 1 $\rightarrow$ 2]{%
    \includegraphics[width=0.22\textwidth]{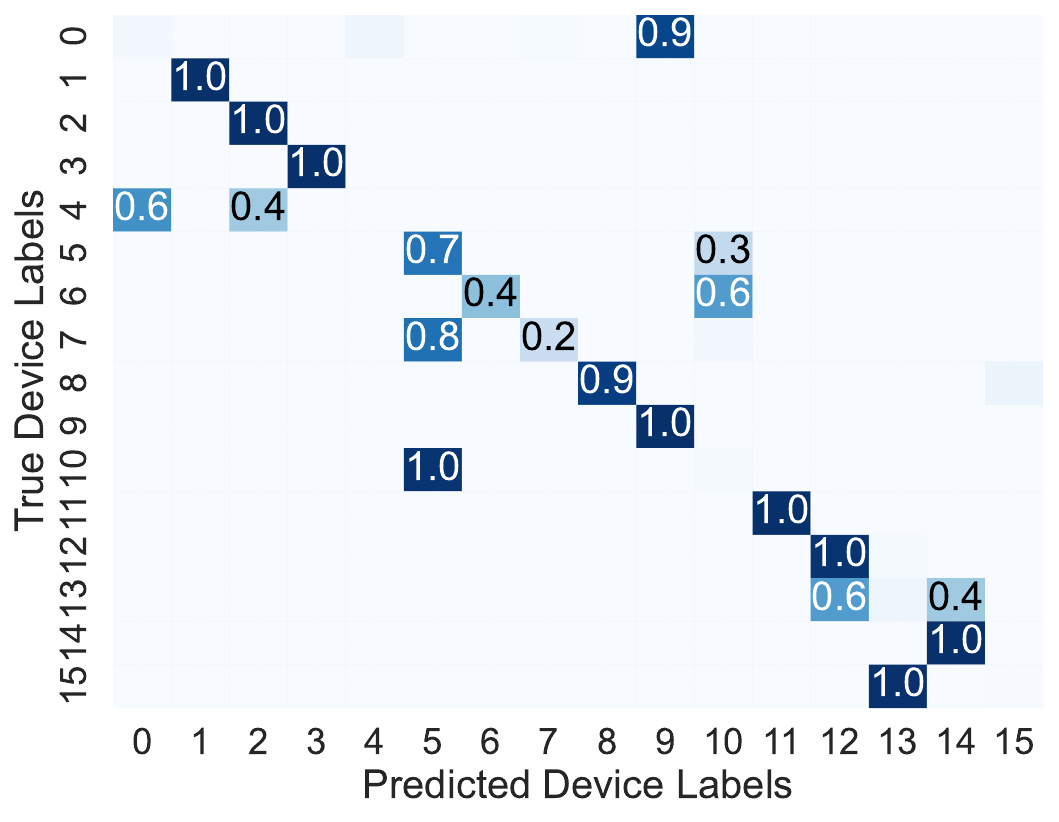}
    \label{fig:expr_cm_ab_wired12x}} \hspace*{-0.1em} 
  \subfloat[AB, Day 2 $\rightarrow$ 1]{%
    \includegraphics[width=0.22\textwidth]{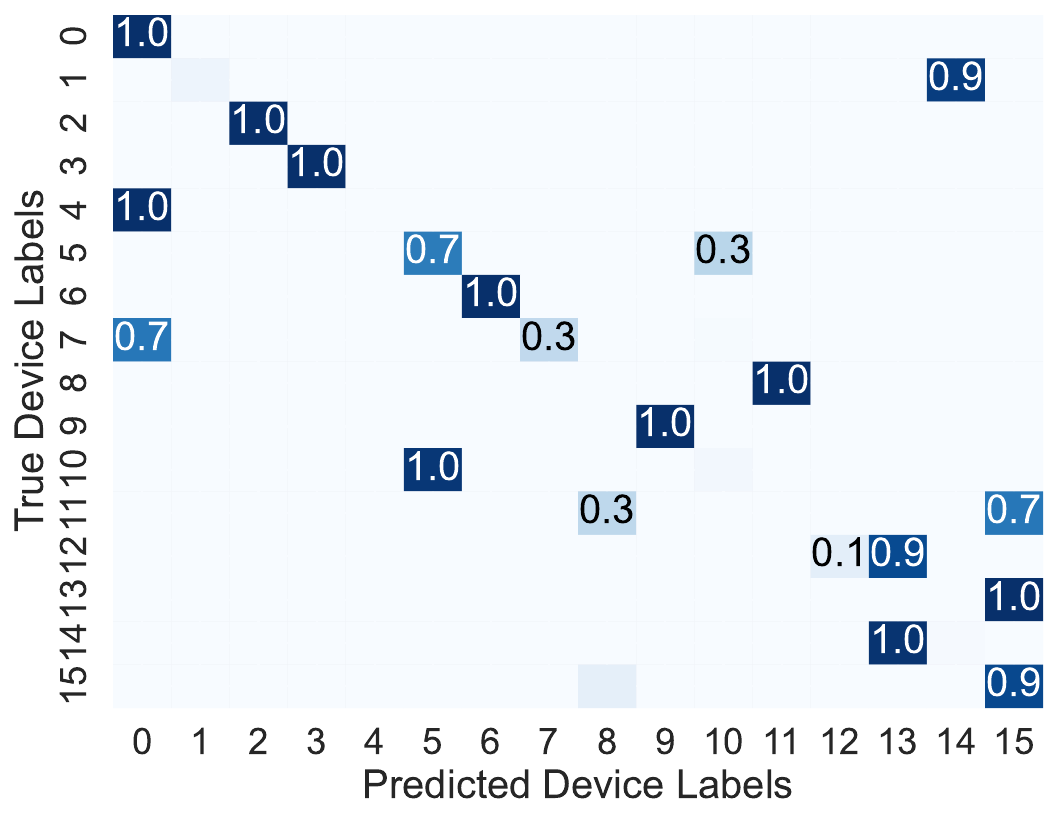}
    \label{fig:expr_cm_ab_wired21x}} \hspace*{-0.1em} 
  \newline
  \subfloat[CTL, Day 1 $\rightarrow$ 2]{%
    \includegraphics[width=0.22\textwidth]{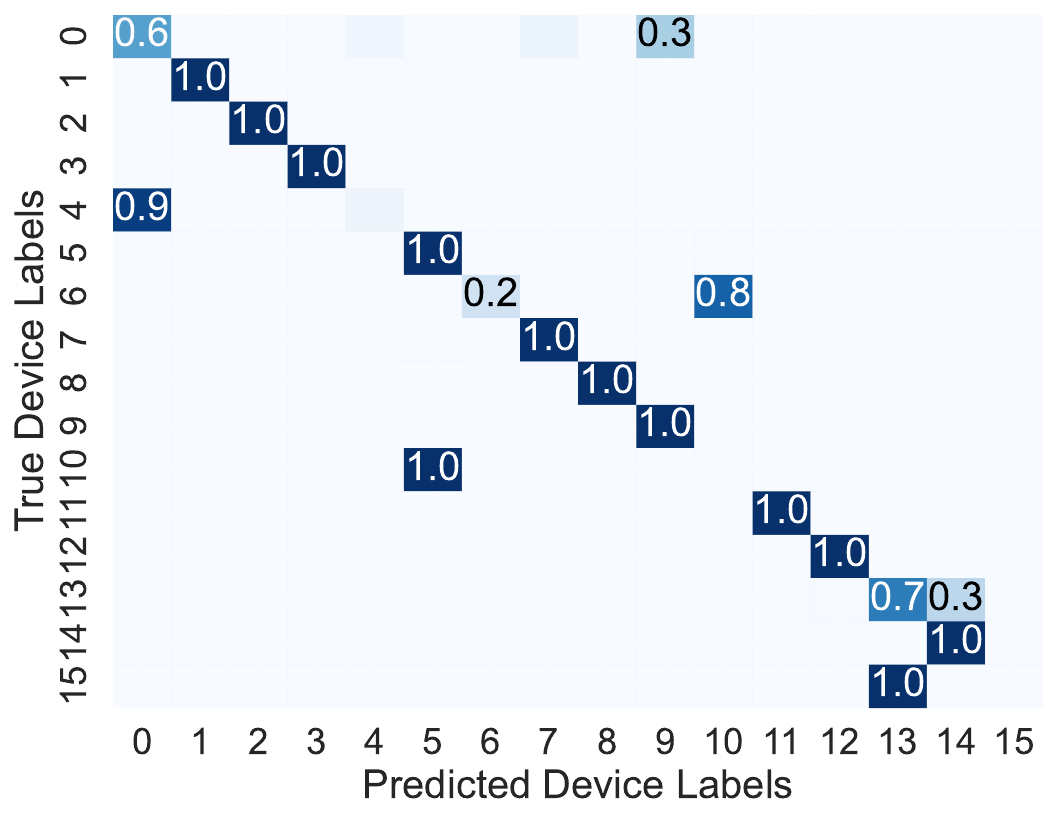}
    \label{fig:expr_cm_ctl_wired12x}} \hspace*{-0.1em} 
  \subfloat[CTL, Day 2 $\rightarrow$ 1]{%
    \includegraphics[width=0.22\textwidth]{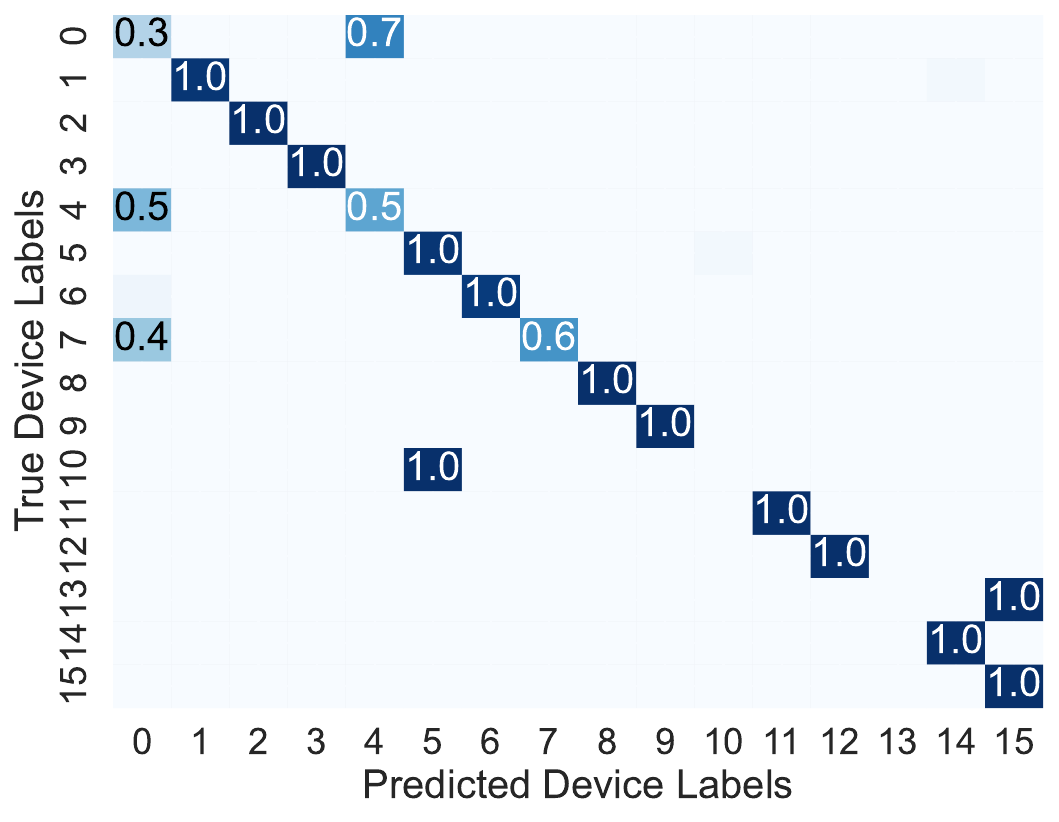}
    \label{fig:expr_cm_ctl_wired21x}} \hspace*{-0.1em}
  \caption{Confusion matrices between day 1 and day 2 on the \textbf{wired} setup for CNN, AB and CTL. Confusion matrices are normalized by row, enabling a clearer visualization of the predicted accuracy distribution across different classes.} 
  \label{fig:expr_cm_wired12x}
\end{figure}


The confusion matrices (Fig.~\ref{fig:expr_cm_wired12x}) show that the CTL model effectively shifts the prediction distribution from off-diagonal to diagonal elements compared to the CNN and AB models. This shift results in fewer misclassifications. For instance, with device 14, the CNN model misclassifies 80\% of the data as device 1 (see Fig.~\ref{fig:expr_cm_cnn_wired12x}), whereas the CTL model correctly classifies 100\% of the data (see Fig.~\ref{fig:expr_cm_ctl_wired12x}). 
Similarly, for device 13, the CNN model misclassifies 80\% as device 14, while the CTL model only misclassifies 30\%. 
Our experimental studies, therefore, indicate that contrastive learning results in a classification model that makes more precise, confident predictions, resulting in a more accurate representation of data patterns. We have also seen similar trends when using the wireless RF data, though results for the wireless data scenario were not included here due to the space limit.


From these confusion matrices, we are also able to identify specific devices that are challenging to classify. Notably, for the wired RF data scenario, some devices (e.g. devices 4 and 10) are invariably misclassified, regardless of the model (CNN, AB, CTL) or the day ("Day 1 $\rightarrow$ 2", "Day 2 $\rightarrow$ 1"). Device 10, in particular, is consistently mistaken for device 5 with high probability in all cases, warranting further investigation.

\section{Conclusion}
\label{sec:conclusion}




This work is the first to apply contrastive learning to the problem of RF fingerprinting under domain shift by constructing positive pairs from the same transmission. Contrastive learning results in large improvements in classification accuracy, largely due to the help of target data, without device labels, in the pre-training stage. Further research, however, is needed to explore several areas. First, investigating the impact of wireless channel impairments, such as fading and mobility, is essential to improving the model's robustness. Second, improving the scalability of contrastive learning is important, especially in large-scale deployments, as the contrastive learning approach requires a significant increase in the amount of data as the number of devices grows. Finally, the confusion matrices from our experiments highlight that certain devices are consistently misclassified, possibly due to hardware issues, and we plan to investigate the cause of the misclassifications.



\bibliographystyle{IEEEtran}
\bibliography{reference}

\begin{thebibliography}{10}
\providecommand{\url}[1]{#1}
\csname url@samestyle\endcsname
\providecommand{\newblock}{\relax}
\providecommand{\bibinfo}[2]{#2}
\providecommand{\BIBentrySTDinterwordspacing}{\spaceskip=0pt\relax}
\providecommand{\BIBentryALTinterwordstretchfactor}{4}
\providecommand{\BIBentryALTinterwordspacing}{\spaceskip=\fontdimen2\font plus
\BIBentryALTinterwordstretchfactor\fontdimen3\font minus
  \fontdimen4\font\relax}
\providecommand{\BIBforeignlanguage}[2]{{%
\expandafter\ifx\csname l@#1\endcsname\relax
\typeout{** WARNING: IEEEtran.bst: No hyphenation pattern has been}%
\typeout{** loaded for the language `#1'. Using the pattern for}%
\typeout{** the default language instead.}%
\else
\language=\csname l@#1\endcsname
\fi
#2}}
\providecommand{\BIBdecl}{\relax}
\BIBdecl

\bibitem{Soltanieh2020ARO}
N.~Soltanieh \emph{et~al.}, ``A review of radio frequency fingerprinting
  techniques,'' \emph{IEEE J. Radio Freq. Identif.}, vol.~4, pp. 222--233,
  2020.

\bibitem{Jian2020DeepLF}
T.~Jian \emph{et~al.}, ``Deep learning for rf fingerprinting: A massive
  experimental study,'' \emph{IEEE Internet of Things Magazine}, vol.~3, pp.
  50--57, 2020.

\bibitem{hamdaoui2023deep}
B.~Hamdaoui, N.~Basha, and K.~Sivanesan, ``Deep learning-enabled zero-touch
  device identification: Mitigating the impact of channel variability through
  {MIMO} diversity,'' \emph{IEEE Communications Magazine}, vol.~61, no.~6, pp.
  80--85, 2023.

\bibitem{elmaghbub2021lora}
A.~Elmaghbub and B.~Hamdaoui, ``{LoRa} device fingerprinting in the wild:
  Disclosing {RF} data-driven fingerprint sensitivity to deployment
  variability,'' \emph{IEEE Access}, vol.~9, pp. 142\,893--142\,909, 2021.

\bibitem{hamdaoui2020deep}
B.~Hamdaoui, A.~Elmaghbub, and S.~Mejri, ``Deep neural network feature designs
  for {RF} data-driven wireless device classification,'' \emph{IEEE Network},
  vol.~35, no.~3, pp. 191--197, 2020.

\bibitem{AlShawabka2020ExposingTF}
A.~Al-Shawabka \emph{et~al.}, ``Exposing the fingerprint: Dissecting the impact
  of the wireless channel on radio fingerprinting,'' \emph{IEEE INFOCOM 2020 -
  IEEE Conference on Computer Communications}, pp. 646--655, 2020.

\bibitem{puppo2023hinova}
L.~Puppo \emph{et~al.}, ``{HiNoVa}: A novel open-set detection method for
  automating {RF} device authentication,'' \emph{arXiv preprint
  arXiv:2305.09594}, 2023.

\bibitem{hamdaoui2022uncovering}
B.~Hamdaoui and A.~Elmaghbub, ``Uncovering the portability limitation of deep
  learning-based wireless device fingerprints,'' \emph{arXiv preprint
  arXiv:2211.07687}, 2022.

\bibitem{Zhao2019OnLI}
H.~Zhao \emph{et~al.}, ``On learning invariant representations for domain
  adaptation,'' in \emph{International Conference on Machine Learning},
  vol.~97, 2019, pp. 7523--7532.

\bibitem{Sun2016DeepCC}
B.~Sun and K.~Saenko, ``Deep coral: Correlation alignment for deep domain
  adaptation,'' in \emph{ECCV Workshops}, 2016.

\bibitem{Ganin2015DomainAdversarialTO}
Y.~Ganin \emph{et~al.}, ``Domain-adversarial training of neural networks,''
  \emph{ArXiv}, vol. abs/1505.07818, 2015.

\bibitem{Bousmalis2016DomainSN}
K.~Bousmalis \emph{et~al.}, ``Domain separation networks,'' in \emph{NeurIPS},
  2016.

\bibitem{Elmaghbub2023ADLIDAD}
A.~Elmaghbub, B.~Hamdaoui, and W.-K. Wong, ``Adl-id: Adversarial
  disentanglement learning for wireless device fingerprinting temporal domain
  adaptation,'' \emph{ArXiv}, vol. abs/2301.12360, 2023.

\bibitem{Wilson2018ASO}
G.~Wilson and D.~J. Cook, ``A survey of unsupervised deep domain adaptation,''
  \emph{ACM TIST}, vol.~11, pp. 1 -- 46, 2018.

\bibitem{wang2021specific}
J.~Wang \emph{et~al.}, ``Specific emitter identification based on deep
  adversarial domain adaptation,'' in \emph{2021 4th International Conference
  on Information Communication and Signal Processing (ICICSP)}.\hskip 1em plus
  0.5em minus 0.4em\relax IEEE, 2021, pp. 104--109.

\bibitem{wang2022specific}
T.~Wang \emph{et~al.}, ``Specific emitter identification based on the
  multi-discrepancy deep adaptation network,'' \emph{IET Radar, Sonar \&
  Navigation}, vol.~16, no.~12, pp. 2079--2088, 2022.

\bibitem{Noroozi2016UnsupervisedLO}
M.~Noroozi and P.~Favaro, in \emph{Computer Vision -- ECCV 2016}, B.~Leibe
  \emph{et~al.}, Eds.\hskip 1em plus 0.5em minus 0.4em\relax Springer
  International Publishing, 2016, pp. 69--84.

\bibitem{He2019MomentumCF}
K.~He \emph{et~al.}, ``Momentum contrast for unsupervised visual representation
  learning,'' \emph{2020 IEEE/CVF CVPR}, pp. 9726--9735, 2019.

\bibitem{Chen2020ImprovedBW}
X.~Chen \emph{et~al.}, ``Improved baselines with momentum contrastive
  learning,'' \emph{ArXiv}, vol. abs/2003.04297, 2020.

\bibitem{Chen2021AnES}
X.~Chen, S.~Xie, and K.~He, ``An empirical study of training self-supervised
  vision transformers,'' \emph{2021 IEEE/CVF ICCV}, pp. 9620--9629, 2021.

\bibitem{Balestriero2023ACO}
R.~Balestriero \emph{et~al.}, ``A cookbook of self-supervised learning,''
  \emph{ArXiv}, vol. abs/2304.12210, 2023.

\bibitem{Grill2020BootstrapYO}
J.-B. Grill \emph{et~al.}, ``Bootstrap your own latent: A new approach to
  self-supervised learning,'' \emph{ArXiv}, vol. abs/2006.07733, 2020.

\bibitem{Chen2020ExploringSS}
X.~Chen and K.~He, ``Exploring simple siamese representation learning,''
  \emph{2021 IEEE/CVF CVPR}, pp. 15\,745--15\,753, 2020.

\bibitem{ozyurt2023contrastive}
Y.~Ozyurt, S.~Feuerriegel, and C.~Zhang, ``Contrastive learning for
  unsupervised domain adaptation of time series,'' in \emph{The Eleventh ICLR},
  2023.

\bibitem{Thota2021ContrastiveDA}
M.~Thota and G.~Leontidis, ``Contrastive domain adaptation,'' \emph{2021
  IEEE/CVF Conference on Computer Vision and Pattern Recognition Workshops
  (CVPRW)}, pp. 2209--2218, 2021.

\bibitem{hamdaoui2022deep}
B.~Hamdaoui and A.~Elmaghbub, ``Deep-learning-based device fingerprinting for
  increased lora-iot security: Sensitivity to network deployment changes,''
  \emph{IEEE network}, vol.~36, no.~3, pp. 204--210, 2022.

\bibitem{johnson2023domain}
B.~Johnson and B.~Hamdaoui, ``Domain-adaptive device fingerprints for network
  access authentication through multifractal dimension representation,''
  \emph{arXiv preprint arXiv:2308.07925}, 2023.

\bibitem{elmaghbub2023eps}
A.~Elmaghbub and B.~Hamdaoui, ``{EPS}: Distinguishable {IQ} data representation
  for domain-adaptation learning of device fingerprints,'' \emph{arXiv preprint
  arXiv:2308.04467}, 2023.

\bibitem{He2015DeepRL}
K.~He \emph{et~al.}, ``Deep residual learning for image recognition,''
  \emph{2016 IEEE CVPR}, pp. 770--778, 2015.

\bibitem{Um2017DataAO}
T.~T. Um \emph{et~al.}, ``Data augmentation of wearable sensor data for
  parkinson’s disease monitoring using convolutional neural networks,''
  \emph{Proceedings of the 19th ACM International Conference on Multimodal
  Interaction}, 2017.

\bibitem{Frosst2019AnalyzingAI}
N.~Frosst, N.~Papernot, and G.~E. Hinton, ``Analyzing and improving
  representations with the soft nearest neighbor loss,'' in \emph{International
  Conference on Machine Learning}, 2019.

\end{thebibliography}

%

\end{document}